\documentclass{article}

\PassOptionsToPackage{numbers,compress}{natbib}
\usepackage[preprint]{neurips_2026}
\usepackage[utf8]{inputenc}
\usepackage[T1]{fontenc}
\usepackage{hyperref}
\usepackage{url}
\usepackage{booktabs}
\usepackage{amsfonts}
\usepackage{amsmath,amssymb}
\usepackage{nicefrac}
\usepackage{microtype}
\usepackage{xcolor}
\usepackage{graphicx}
\title{Benchmarking System One Models for Time-Series Decision Making: A Case Study of Jev}
\author{%
  Jianye Wang\quad  Kexin Zhang\quad  Jiayin Xiong\quad  Hao Xue\thanks{Corresponding author.} \\
  The Hong Kong University of Science and Technology (Guangzhou)
}

\begin{document}
\maketitle

\begin{abstract}
Time-series tasks require both accurate predictions and fast decisions. Advances in large models have introduced decision-oriented models that directly output answers and their probabilities, which we call Structured System 1 models. Their applicability to time-series tasks remains underexplored. Using Jev 1.13.0 as a case study, we evaluate time-series classification, event type prediction, oil temperature direction and stock movement prediction, converting continuous forecasting targets into discrete decisions to fit the model interface. We establish a unified evaluation framework using accuracy and AUC. Results show stronger performance on selected explicit temporal patterns and event type prediction, but near-chance discrimination on many numerical sequence classification tasks and stock movement prediction. Some datasets also reveal a gap between ranking quality and final decision accuracy. These findings motivate improvements in temporal representation, domain adaptation, and decision-score alignment. Standardized sample construction, model calls, and result recording support extending the evaluation to future models.
\end{abstract}

\section{Introduction}

Time series record how a system changes over time and support applications across diverse domains: wearable sensors enable human activity recognition \citep{reiss2012pamap}, physiological measurements support patient assessment and deterioration prediction \citep{harutyunyan2019clinical}, operational signals enable equipment fault detection and diagnosis \citep{lessmeier2016bearing}, and financial observations support market movement prediction \citep{patel2015stock}. These applications require time-series analysis to balance accuracy and processing speed. Traditional machine learning methods, including support vector machines, random forests, and XGBoost \citep{cortes1995svm,breiman2001randomforests,chen2016xgboost}, are typically trained using task-specific features, and their predictive performance depends on feature representations and training data.

Recent advances in large language models and time-series foundation models have introduced new approaches to time-series analysis. Time-LLM maps temporal patches into the representation space of a frozen language model, Chronos quantizes time-series values for pretraining across domains, and Chronos-2 further supports zero-shot multivariate and covariate-informed forecasting \citep{jin2024timellm,ansari2024chronos,ansari2025chronos2}. Some studies use natural-language instructions to guide models in making time-series predictions \citep{liu2025system12tsf}.

\begin{figure}[t]
    \centering
    \includegraphics[width=\linewidth]{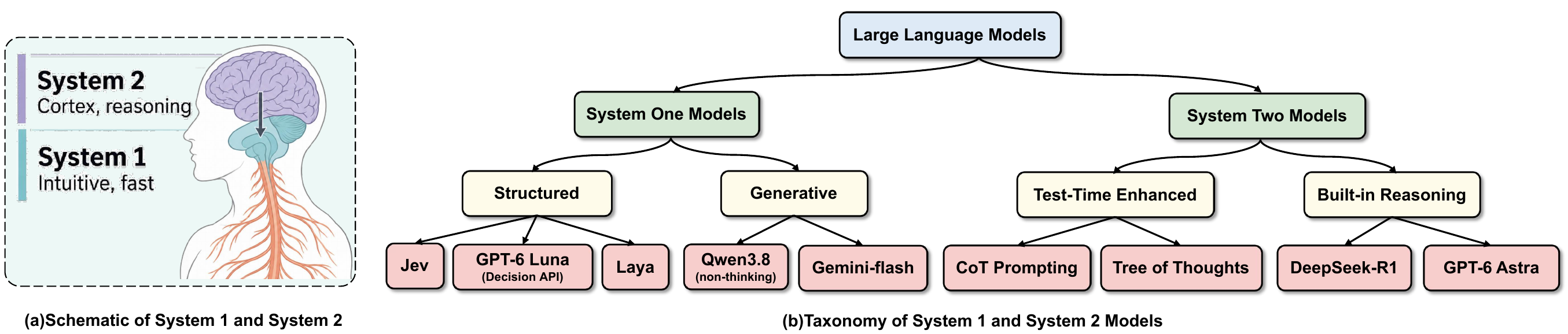}
    \caption{(a) Schematic of System 1 and System 2, illustrating intuitive judgments and deliberative reasoning in dual-process theory. (b) Model taxonomy under our operational definitions: System 1 provides direct answers through structured outputs or text generation, while System 2 uses intermediate reasoning through test-time enhancement or built-in reasoning.}
    \label{fig:model_taxonomy}
\end{figure}

Although complex temporal dependencies may require stronger modeling and reasoning capabilities, many time-series tasks can be naturally formulated as direct decision problems rather than continuous-valued forecasting. A unified, direct, and structured decision-making approach also offers potential advantages in computational efficiency and reduced generation overhead. These considerations motivate evaluating whether Structured System 1 can effectively interpret time-series information and produce reliable decisions. Inspired by dual-process theories in psychology, related studies use System 1 to describe fast, direct judgments and System 2 to describe deliberative reasoning involving multiple steps \citep{evans2013dualprocess,yu2024distilling}, as shown in Figure~\ref{fig:model_taxonomy}(a).

Existing studies differ in how they distinguish System 1 from System 2. \emph{Distilling System 2 into System 1} categorizes Chain-of-Thought (CoT) and other methods that generate intermediate reasoning steps as System 2 \citep{yu2024distilling}. The time-series forecasting benchmark ReC4TS instead treats CoT, self-consistency, and self-correction as ``System 1 reasoning enhancements,'' while categorizing models that undergo dedicated reasoning post-training as System 2 \citep{liu2025system12tsf}.

This paper adopts an operational distinction based on the inference procedure. As shown in Figure~\ref{fig:model_taxonomy}(b), System 1 produces final answers without generating an intermediate reasoning sequence and is divided into \emph{Structured System 1} and \emph{Generative System 1} according to the output mechanism. The former directly computes decisions, while the latter generates answers through text decoding, with implications for response speed and output stability. System 2 uses additional inference-time computation for intermediate derivation, search, or reflection through \emph{Built-in Reasoning} or \emph{Test-Time Enhanced} approaches, which can be combined. Existing time-series research has focused primarily on System 2 and Generative System 1, with fewer systematic evaluations of Structured System 1.

Structured System 1 models score predefined alternatives, labels, or ordered levels conditioned on a state, a natural-language question, and candidate answers, returning decisions and probabilities through a unified interface. This output format supports time-series classification, anomaly detection, and discretized forecasting. A typical implementation uses an encoder to obtain contextual representations and a classification or scoring head to map them into the target output space, reducing sequential decoding and answer-parsing overhead. BERT combines a pretrained encoder with task-specific output layers for classification and natural language inference, while DistilBERT explores smaller and faster language representation models, providing a foundation for this approach \citep{devlin2019bert,sanh2019distilbert}.

Jev is a System One model developed by TypeSafe AI, characterized by low latency and low API cost \citep{typesafe2026jev}. Independent research has evaluated its decision quality and invocation cost on text classification, natural language inference, reading comprehension, and content moderation \citep{deusser2026jev}, while its performance on time-series tasks awaits systematic evaluation. We use Jev as a case study, organizing temporal observations into a unified state-and-question representation and evaluating decision quality using accuracy (ACC) and the area under the receiver operating characteristic curve (AUC) to assess the applicability of Structured System 1 across time-series tasks.

Our contributions are threefold:
\begin{itemize}
    \item[--] We define the task scope and decision mappings for
    time-series classification, anomaly detection, and adapted
    forecasting, mapping continuous future values to explicit
    decision targets.

    \item[--] We evaluate Jev 1.13.0 through binary and multiclass
    decision interfaces, reporting ACC and AUC and analyzing
    performance across datasets and task types.

    \item[--] We establish an extensible evaluation framework
    that standardizes sample construction, model calls, and
    result recording, enabling subsequent Structured System 1
    models to be evaluated under the same task definitions and
    evaluation protocols.
\end{itemize}

\section{Problem formulation}

\subsection{Time-series inputs and decision outputs}

Following established time-series definitions \citep{jin2024gnnsurvey,jin2026survey}, let
\begin{equation}
    \mathbf X=(\mathbf x_1,\ldots,\mathbf x_T)\in\mathbb R^{T\times D},
    \qquad \mathbf x_t\in\mathbb R^D,
\end{equation}
where $T$ is the number of ordered observations, $D$ is the number of channels, and $\mathbf x_t$ contains the values at time $t$. A univariate series is the special case $D=1$. For a decision at time $t$, an observed window of length $L$ is
\begin{equation}
    \mathbf X^{(t)}=\mathbf X_{t-L+1:t}
    =(\mathbf x_{t-L+1},\ldots,\mathbf x_t)\in\mathbb R^{L\times D},
    \qquad L\leq t\leq T.
\end{equation}
The right endpoint is $t$ and values after $t$ cannot be used in the input to a future-oriented decision.

Some observations are events at irregular times. We write their history as
\begin{equation}
    \mathcal H_t=\{(\tau_i,m_i):\tau_i\leq t,\ i=1,\ldots,n_t\},
\end{equation}
where $\tau_i$ is the event time, $m_i$ is its categorical mark, and $n_t$ is the number of observed events by time $t$. This representation retains the event times instead of imposing a regular sampling interval. We use $\mathcal I$ for the task-specific input, whether a complete series, a window, or an event history.

Let $s=\sigma(\mathcal I)$ be the state supplied to the model, with $\sigma$ denoting a fixed representation procedure. Given a question $q$ and a candidate set $\mathcal Y=\{1,\ldots,K\}$, a decision model returns
\begin{equation}
    \mathbf p_\theta(s,q,\mathcal Y)
    =\bigl(p_\theta(k\mid s,q,\mathcal Y)\bigr)_{k=1}^K,
    \qquad \sum_{k=1}^K p_\theta(k\mid s,q,\mathcal Y)=1,
    \label{eq:decision}
\end{equation}
with nonnegative components. Here $\theta$ denotes model parameters and $K$ is the number of admissible decisions. The predicted answer is $\hat y=\arg\max_{k\in\mathcal Y}p_\theta(k\mid s,q,\mathcal Y)$. The three tasks below differ in what is observed, what the label denotes, and when it becomes known.

\subsection{Time-series classification}

For sample $i$, let $\mathbf X^{(i)}\in\mathbb R^{T_i\times D}$ be an observed series of length $T_i$ and let $y_i\in\{1,\ldots,K\}$ be its class. The goal is to estimate
\begin{equation}
    p_\theta\!\left(y_i=k\mid\sigma(\mathbf X^{(i)}),q,\mathcal Y\right),
    \qquad
    \hat y_i=\arg\max_{k\in\mathcal Y}
    p_\theta\!\left(y_i=k\mid\sigma(\mathbf X^{(i)}),q,\mathcal Y\right).
\end{equation}
The index $i$ identifies a sample, $T_i$ its length, and $D$ the number of channels within its dataset. A segment cut from a continuous activity record can itself be a classification sample. For $K=2$ the task is binary, for $K>2$ it is multiclass. The label describes the observed segment rather than an unobserved future.

\subsection{Time-series anomaly detection}

Let $a_t\in\{0,1\}$ indicate whether time $t$ is anomalous, with $1$ denoting an anomaly. Using the historical window ending at $t$, define the anomaly score
\begin{equation}
    r_t=p_\theta\!\left(a_t=1\mid\sigma(\mathbf X^{(t)}),q,\{0,1\}\right),
    \qquad \hat a_t=\mathbb I[r_t\geq\eta],
    \label{eq:anomaly}
\end{equation}
where $\eta\in[0,1]$ is a prespecified decision threshold and $\mathbb I[\cdot]$ is the indicator function. If anomalies are annotated as intervals $[u_j,v_j]$, the corresponding point label is $a_t=\mathbb I[t\in\bigcup_j[u_j,v_j]]$. The window supplies evidence, whereas its right endpoint is the object being labeled. This distinction prevents the label ``the window contains an anomaly'' from silently replacing the label ``time $t$ is anomalous.'' The continuous score $r_t$ can also be used to compute threshold-independent ranking measures.

\subsection{Decision-oriented time-series forecasting}

Conventional forecasting estimates future continuous observations. Let $H$ be the forecast horizon and $D_y$ the number of forecast target channels. With $\mathbf y_{t+h}\in\mathbb R^{D_y}$, the future target is
\begin{equation}
    \mathbf Y^{(t)}=(\mathbf y_{t+1},\ldots,\mathbf y_{t+H})
    \in\mathbb R^{H\times D_y},
    \qquad
    \widehat{\mathbf Y}^{(t)}=f_\theta(\mathbf X^{(t)}),
    \label{eq:forecast}
\end{equation}
where $1\leq h\leq H$ denotes a future step and $f_\theta$ is a numeric forecasting model. For example, squared-error training minimizes a sum of $\|\mathbf Y^{(t)}-f_\theta(\mathbf X^{(t)})\|_F^2$ over training examples, with $\|\cdot\|_F$ denoting the Frobenius norm.

Our reformulation fixes a label map $\phi$ before evaluating the model:
\begin{equation}
    z_t=\phi(\mathbf Y^{(t)};\lambda)\in\{1,\ldots,K\},
    \quad
    \hat z_t=\arg\max_{k\in\mathcal Y}
    p_\theta\!\left(z_t=k\mid\sigma(\mathbf X^{(t)}),q,\mathcal Y\right).
    \label{eq:reformulation}
\end{equation}
Here $\lambda$ collects task parameters, such as a horizon or threshold. The future values generate the ground-truth label $z_t$, but are never included in the input state. The model directly answers a question about an event in the future.

For a concrete example, let $P_t>0$ be a stock's closing price at time $t$, and define the $h$-step return, for $h\leq H$, by
\begin{equation}
    R_{t,h}=\frac{P_{t+h}-P_t}{P_t}.
\end{equation}
The binary upward-movement target is
\begin{equation}
    z_t=\mathbb I[R_{t,h}>0]=\mathbb I[P_{t+h}>P_t].
    \label{eq:stockbinary}
\end{equation}
The equivalence follows because dividing by the positive $P_t$ does not change the sign of $P_{t+h}-P_t$. The requested quantity is thus $\Pr(P_{t+h}>P_t\mid\mathbf X^{(t)})$, rather than a point estimate of $P_{t+h}$. A three-class version, with a fixed tolerance $\delta>0$, uses
\begin{equation}
    z_t=\begin{cases}
    \text{down}, & R_{t,h}<-\delta,\\
    \text{flat}, & -\delta\leq R_{t,h}\leq\delta,\\
    \text{up}, & R_{t,h}>\delta.
    \end{cases}
\end{equation}
The threshold $\delta$ determines the width of the flat class and must remain fixed across systems being compared. Stock movement prediction has previously been posed as a directional task \citep{xu2018stock}, the example here illustrates the general conversion of a continuous forecast target into an explicit decision event.

The reformulation changes the statistical object being estimated. For a scalar future price $U=P_{t+h}$ and a squared-error point forecast $c$, differentiating its conditional loss gives
\begin{equation}
    \frac{\partial}{\partial c}
    \mathbb E[(U-c)^2\mid\mathbf X^{(t)}]
    =2\bigl(c-\mathbb E[U\mid\mathbf X^{(t)}]\bigr),
    \quad
    c^*=\mathbb E[U\mid\mathbf X^{(t)}].
\end{equation}
In contrast, the binary decision depends on $\Pr(U>P_t\mid\mathbf X^{(t)})$. A conditional mean and a probability mass above a threshold summarize different aspects of the future distribution. The same historical window can therefore lead to different optimal numerical and decision outputs.

\subsection{Mapping tasks to typed answers}

Equation~\eqref{eq:decision} gives a common output form. Sequence classification uses the class of the observed sample, anomaly detection uses the status of a specified endpoint, and decision-oriented forecasting uses a future-derived class. Binary questions are evaluated through Jev's Noul primitive, which returns the probability that a statement is true. Choice selects from a fixed candidate set and returns option probabilities, supporting both binary and multiclass decisions. The label construction and the evaluation unit remain task-specific even though the external decision format is shared.

\begin{figure}[htbp]
    \centering
    \includegraphics[width=\linewidth]{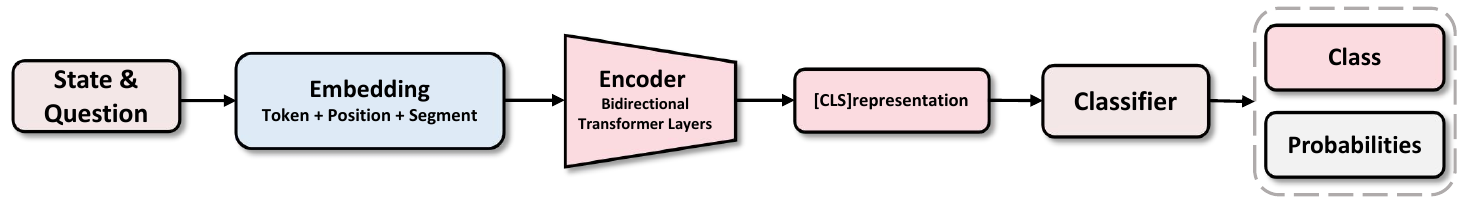}
\caption{One possible architecture for a Structured System 1 model, illustrated using BERT. Embedding, encoding, and classification map the input to class probabilities and a predicted class.}
    \label{fig:model_method}
\end{figure}

\section{Method}

Jev is a commercial model whose full internal architecture is not publicly documented~\citep{typesafe2026jev}. Figure~\ref{fig:model_method} illustrates one possible architecture for Structured System 1 decision making using a BERT-based classifier~\citep{devlin2019bert}. We use this example to explain the main computational stages and evaluate Jev through its public decision interface.

\subsection{Input Embedding}

Given an observed time series $\mathcal I$, a representation procedure constructs the input state $s=\sigma(\mathcal I)$. The state records observations chronologically and can include channel names and timestamps. Together with the task question $q$, it is serialized into a token sequence:
\begin{equation}
    \mathbf u
    =\operatorname{Tokenize}
    \left(
    [\mathrm{CLS}]\,s\,[\mathrm{SEP}]\,q\,[\mathrm{SEP}]
    \right).
\end{equation}
Here, $[\mathrm{CLS}]$ is the classification token and $[\mathrm{SEP}]$ separates the state and question. Numerical observations are represented through the tokens produced by the tokenizer.

BERT combines token, position, and segment embeddings to construct the initial representation of token $u_j$:
\begin{equation}
    \mathbf h_j^{(0)}
    =
    \mathbf e_{\mathrm{tok}}(u_j)
    +\mathbf e_{\mathrm{pos}}(j)
    +\mathbf e_{\mathrm{seg}}(a_j),
\end{equation}
where $a_j$ identifies whether the token belongs to the state or question. For $N$ tokens and hidden dimension $d$, these vectors form
$\mathbf H^{(0)}\in\mathbb R^{N\times d}$.
Token embeddings represent the input content, position embeddings distinguish token locations, and segment embeddings distinguish the two input components~\citep{devlin2019bert}.

\subsection{Contextual Encoder}

The embedding sequence passes through a stack of bidirectional Transformer encoder layers:
\begin{equation}
    \mathbf H^{(\ell)}
    =
    \operatorname{Encoder}_{\ell}
    \left(\mathbf H^{(\ell-1)}\right),
    \qquad \ell=1,\ldots,L_{\mathrm{enc}},
\end{equation}
where $L_{\mathrm{enc}}$ is the number of encoder layers. Each layer uses multi-head self-attention and a feed-forward network, together with residual connections and layer normalization \citep{vaswani2017attention}.

For one attention head, the input representations are projected into queries, keys, and values:
\begin{equation}
    Q=\mathbf H W_Q,\qquad
    K=\mathbf H W_K,\qquad
    V=\mathbf H W_V.
\end{equation}
The attention output is
\begin{equation}
    \operatorname{Attention}(Q,K,V)
    =
    \operatorname{softmax}
    \left(\frac{QK^\top}{\sqrt{d_k}}\right)V,
\end{equation}
where $d_k$ is the key dimension. This operation allows each token to incorporate information from other tokens in the observed input. In the time-series setting, it provides a mechanism for relating observations at different positions and conditioning their interpretation on the task question.

The final classification-token representation,
\begin{equation}
    \mathbf h_{\mathrm{CLS}}
    =
    \mathbf H^{(L_{\mathrm{enc}})}_{\mathrm{CLS}}
    \in\mathbb R^d,
\end{equation}
provides a contextual representation for the decision head. Bidirectional encoding operates over the supplied observations, which contain only the history available for the task.

\subsection{Classification and Decision Probabilities}

For a fixed answer set $\mathcal Y=\{1,\ldots,K\}$, a linear classification head maps the contextual representation to class logits:
\begin{equation}
    \boldsymbol\ell
    =
    W\mathbf h_{\mathrm{CLS}}+\mathbf b
    \in\mathbb R^K,
\end{equation}
where $W\in\mathbb R^{K\times d}$ and
$\mathbf b\in\mathbb R^K$ are learned parameters. A softmax converts these logits into a probability distribution:
\begin{equation}
    p_\theta(y=k\mid s,q,\mathcal Y)
    =
    \frac{\exp(\ell_k)}
    {\sum_{j=1}^{K}\exp(\ell_j)},
    \qquad
    \hat y=\arg\max_{k\in\mathcal Y}
    p_\theta(y=k\mid s,q,\mathcal Y).
\end{equation}

For binary decisions, a single logit
$v=\mathbf w^\top\mathbf h_{\mathrm{CLS}}+b$
can instead be converted into the positive-class probability:
\begin{equation}
    p_\theta(y=1\mid s,q)=\frac{1}{1+e^{-v}},
    \qquad
    p_\theta(y=0\mid s,q)=1-p_\theta(y=1\mid s,q).
\end{equation}

The fixed-label head illustrates standard BERT classification. To accommodate changing candidate sets, an alternative is to encode each candidate together with the state and question, apply a shared scoring head, and normalize the resulting scores across candidates. This makes the output depend explicitly on the supplied alternatives.

In both cases, the encoder representations are mapped directly to decision probabilities without generating an answer token by token. The predicted class supports ACC evaluation, while the probabilities support AUC evaluation. In our Jev experiments, these outputs are obtained through Noul for binary questions and Choice for multiclass questions.

\section{Experiments}

\subsection{Datasets And Evaluation Metrics}

This study selects datasets suitable for evaluating System 1 models, covering time series classification, event type prediction, time series anomaly detection, and forecasting. These datasets include univariate and multivariate numerical sequences, as well as irregular event sequences consisting of timestamps and event types. We evaluate Jev on these datasets using accuracy (ACC) and the area under the receiver operating characteristic curve (AUC).

\paragraph{(1) Time series classification: UCR and UEA archives.}

The UCR Archive is a representative collection of datasets for univariate time series classification. Its 2018 release contains 128 datasets, covering sensor recordings, physiological signals such as electrocardiograms, motion data, and sequences derived from image outlines~\cite{dau2019ucr}. Each sample consists of a univariate sequence and its class label. The number of samples, sequence lengths, and numbers of classes vary substantially across datasets. The task is therefore \textbf{univariate time series classification}: the model reads an entire sequence and predicts its class, under either a binary or multiclass setting. This study uses the archive to examine Jev's ability to recognize temporal patterns in a single channel.

The UEA Archive focuses on \textbf{multivariate time series classification}. Its 2018 release contains 30 datasets covering activity recognition, motion recognition, physiological signals, and other settings involving multiple observation channels~\cite{bagnall2018uea}. Whereas each time step in UCR contains a single observation, each UEA sample consists of multiple channels evolving over time and is associated with a sequence-level class label. The model must identify temporal changes within individual channels and exploit relationships between channels. For example, activity recognition may require combining recordings from several motion sensors to distinguish actions with similar local patterns. UCR and UEA thus provide sequence classification evaluations under univariate and multivariate input settings, respectively.

\paragraph{(2) Event type prediction: EasyTPP.}

EasyTPP is a benchmarking framework for temporal point processes that provides event sequence datasets, model implementations, and standardized evaluation procedures~\cite{xue2024easytpp}. The five real-world datasets used in its original paper are Amazon, Retweet, Taxi, Taobao, and StackOverflow. They contain user reviews, social reposts, taxi pick-up and drop-off events, e-commerce clicks, and user badge awards, respectively. Each sequence consists of events ordered by occurrence time, with each event containing a timestamp and a discrete event type. The intervals between events are generally irregular.



\paragraph{(3) Direction prediction: ETT and stock market data.}

The Electricity Transformer Temperature (ETT) dataset contains transformer monitoring records used in the Informer study~\cite{zhou2021informer}. Its ETT-small collection includes ETTh1, ETTh2, ETTm1, and ETTm2, with hourly or 15-minute observations of six electrical load variables and oil temperature (OT). The stock market data comprise daily records for AAPL, BA, and CAT released by Al Ridhawi et al.~\cite{alridhawi2026ohlcv}. These datasets cover two forecasting settings: temperature changes in electrical equipment and price movements in financial markets.

This study reformulates continuous-valued forecasting on both datasets as \textbf{binary direction prediction}: the model reads a historical observation window and predicts whether the target value at a specified future horizon exceeds its current value, yielding ``increase'' and ``not increase.'' For ETT, the target is OT at a 24-hour horizon, corresponding to 24 hourly steps or 96 15-minute steps. We evaluate univariate (S) and multivariate (MS) input settings, using OT alone or OT together with the six electrical load variables, respectively. For the stock data, the target is the closing price at a fixed number of trading steps ahead. Together, these datasets examine Jev's ability to use temporal observations to anticipate directional changes, providing information relevant to transformer operational monitoring and directional investment decisions.

\subsection{Implementation Details}

Each sample is represented by an input state, a task question, candidate answers, and a ground-truth label. Observations are arranged chronologically using consistent input settings within each experiment.

We evaluate Jev 1.13.0 with a fixed question template for each task. Binary tasks use Noul to estimate the probability of the positive event, while multiclass tasks use Choice to return a selected answer and candidate probabilities. Predictions and probabilities are recorded alongside ground-truth labels for evaluation.

\begin{table}[t]
\centering
\caption{Overall results, averaged equally across datasets with available scores.}
\label{tab:results_overall}
\small
\begin{tabular}{lrrrr}
\hline
Dataset group & Number & Mean ACC & Mean AUC & Median AUC \\
\hline
UCR binary     & 41 & 0.5411 & 0.5341 & 0.5277 \\
UCR multiclass & 86 & 0.1874 & 0.5187 & 0.5047 \\
UEA            & 17 & 0.1566 & 0.5172 & 0.4986 \\
EasyTPP        &  5 & 0.5127 & 0.7028 & 0.6684 \\
 ETT& 8& 0.5137& 0.5900&0.5912\\
Stocks         &  3 & 0.5034 & 0.5040 & 0.5042 \\
\hline
\end{tabular}
\end{table}

\begin{table}[t]
\centering
\caption{Selected UCR results used in the analysis.}
\label{tab:results_ucr}
\small
\begin{tabular}{llrr}
\hline
Setting & Dataset & ACC & AUC \\
\hline
Binary & Chinatown          & 0.6793 & 0.6657 \\
       & DodgerLoopWeekend  & 0.7464 & 0.6953 \\
       & DodgerLoopGame     & 0.5942 & 0.6475 \\
       & ItalyPowerDemand   & 0.5598 & 0.6597 \\
       & GunPoint           & 0.5200 & 0.7215 \\
       & Wafer              & 0.6525 & 0.8056 \\
       & ECG200             & 0.5900 & 0.4696 \\
       & ECGFiveDays        & 0.4901 & 0.4434 \\
       & TwoLeadECG         & 0.5004 & 0.4629 \\
\hline
Multiclass & SyntheticControl & 0.8100 & 0.9978 \\
           & SmoothSubspace   & 0.9000 & 0.9815 \\
           & BME              & 0.3933 & 0.7434 \\
           & CBF              & 0.3889 & 0.7019 \\
           & DodgerLoopDay    & 0.2125 & 0.5056 \\
           & TwoPatterns      & 0.1380 & 0.3224 \\
\hline
\end{tabular}
\end{table}

\subsection{Experimental Results and Analysis}

Table~\ref{tab:results_overall} summarizes Jev's performance across the evaluated dataset groups. Dataset-level results supporting the analysis are presented in Tables~\ref{tab:results_ucr}--\ref{tab:results_stock}. (1) Numerical sequence classification shows limited overall discrimination: the mean AUC is 0.5341 for binary UCR tasks, 0.5187 for multiclass UCR tasks, and 0.5172 for UEA tasks. Among the 86 multiclass UCR datasets, 63 have AUC values between 0.45 and 0.55, with a median of 0.5047. (2) Strong performance is concentrated in a small subset. SyntheticControl achieves an ACC of 0.8100 and an AUC of 0.9978, while SmoothSubspace achieves 0.9000 and 0.9815. Excluding these two datasets reduces the mean multiclass UCR AUC to 0.5075. These results show that success on selected temporal patterns has not translated into consistent performance across the archive.

The dataset-level results reveal differences associated with temporal structure, target granularity, and signal type. (1) Several stronger results involve recognizable trends or shapes. SyntheticControl contains classes defined by cycles, trends, and level shifts, while SmoothSubspace distinguishes the position of a continuous segment~\citep{dau2019ucr}. BME and CBF also achieve AUC values of 0.7434 and 0.7019. Chinatown and ItalyPowerDemand reach 0.6657 and 0.6597, suggesting some ability to recognize patterns in everyday activity and demand. However, Wafer reaches 0.8056 despite its specialized industrial setting, whereas the synthetic TwoPatterns dataset reaches only 0.3224. The results therefore suggest that accessibility of the discriminative pattern matters alongside domain familiarity. (2) Target granularity also affects performance: DodgerLoopWeekend achieves an AUC of 0.6953 when distinguishing weekdays from weekends, compared with 0.5056 for the seven-class DodgerLoopDay task. This contrast is consistent with stronger recognition of broad activity differences than fine distinctions within the same application domain. (3) Physiological signals remain difficult. ECG200, ECGFiveDays, and TwoLeadECG have AUC values of 0.4696, 0.4434, and 0.4629, while the EEG dataset FingerMovements reaches 0.4138. Limited domain-specific knowledge and insufficient representation of discriminative signal features are possible explanations. (4) Some scores retain useful ranking information despite weak final decisions. GunPoint achieves an AUC of 0.7215 with an ACC of 0.5200, and BasicMotions achieves an AUC of 0.7671 with an ACC of 0.3000. This gap motivates examining binary thresholds and the comparability of multiclass candidate scores.

\begin{table}[t]
\centering
\caption{Selected UEA results used in the analysis.}
\label{tab:results_uea}
\small
\begin{tabular}{lrr}
\hline
Dataset & ACC & AUC \\
\hline
BasicMotions       & 0.3000 & 0.7671 \\
Epilepsy           & 0.2971 & 0.6622 \\
PenDigits          & 0.1335 & 0.5998 \\
FingerMovements    & 0.4400 & 0.4138 \\
AtrialFibrillation & 0.3333 & 0.3567 \\
\hline
\end{tabular}
\end{table}

\begin{table}[t]
\centering
\caption{Event type prediction results on EasyTPP datasets.}
\label{tab:results_easytpp}
\small
\begin{tabular}{lrr}
\hline
Dataset & ACC & AUC \\
\hline
Amazon    & 0.2795 & 0.6684 \\
Retweet   & 0.5718 & 0.6466 \\
Stockflow & 0.4037 & 0.6470 \\
Taobao    & 0.6343 & 0.8241 \\
Taxi      & 0.6740 & 0.7282 \\
\hline
Mean      & 0.5127 & 0.7028 \\
\hline
\end{tabular}
\end{table}

\begin{table}[t]
\centering
\caption{Oil temperature direction prediction results.}
\label{tab:results_OT}
\small
\begin{tabular}{lcccc}
\hline
& \multicolumn{2}{c}{S}
& \multicolumn{2}{c}{MS} \\
\cline{2-3} \cline{4-5}
Dataset & ACC & AUC & ACC & AUC \\
\hline
ETTh1 & 0.5126 & 0.6006 & 0.4895 & 0.5691 \\
ETTh2 & 0.5102 & 0.5816 & 0.5837 & 0.5935 \\
ETTm1 & 0.4898 & 0.5889 & 0.4909 & 0.6054 \\
ETTm2 & 0.5098 & 0.5559 & 0.5231 & 0.6253 \\
\hline
Mean  & 0.5056 & 0.5818 & 0.5218 & 0.5983 \\
\hline
\end{tabular}

\vspace{2pt}
\begin{minipage}{\columnwidth}
\footnotesize
\textit{Note:} S denotes univariate input (oil
temperature only), while MS denotes multivariate
input (oil temperature and six electrical load
variables). 
\end{minipage}
\end{table}

\begin{table}[t]
\centering
\caption{Binary stock movement prediction results.}
\label{tab:results_stock}
\small
\begin{tabular}{lrr}
\hline
Stock & ACC & AUC \\
\hline
AAPL & 0.5037 & 0.4950 \\
BA   & 0.5000 & 0.5042 \\
CAT  & 0.5064 & 0.5129 \\
\hline
Mean & 0.5034 & 0.5040 \\
\hline
\end{tabular}
\end{table}

Future-oriented tasks show a further contrast between event categories and numerical movement directions. (1) Event type prediction produces consistently higher AUC values in the reported experiments. Across the five EasyTPP datasets, AUC ranges from 0.6466 to 0.8241. Taobao achieves an ACC of 0.6343 and an AUC of 0.8241, while Taxi achieves 0.6740 and 0.7282. These results suggest that discrete event histories provide a useful input form for Jev, potentially because event categories expose predictive relationships more directly than raw numerical sequences. (2) Stock movement prediction remains close to chance-level discrimination. AAPL, BA, and CAT achieve AUC values of 0.4950, 0.5042, and 0.5129, respectively, with a mean ACC of 0.5034. Thus, converting continuous price forecasting into binary movement prediction makes the task compatible with the decision interface but yields little predictive discrimination in these experiments. Oil temperature direction prediction also shows modest discriminative ability, with a mean ACC of 0.5137 and a mean AUC of 0.5900 across the eight ETT settings (Table~\ref{tab:results_OT}). Multivariate input achieves higher AUC on ETTh2, ETTm1, and ETTm2, while univariate input performs better on ETTh1. However, ACC remains close to 0.5 in most settings, suggesting a gap between probability ranking and final classification decisions. Together, the results identify discrete event prediction and selected explicit temporal patterns as the stronger settings, while numerical direction prediction and specialized signal classification remain challenging.

The results suggest that future System 1 models should extend their recognition of explicit trends and simple shapes to finer temporal patterns and cross-channel relationships. Strong performance on SyntheticControl and SmoothSubspace contrasts with weaker results on physiological signals and most evaluated UEA datasets, motivating broader domain-specific training data and improved temporal and channel representations. These developments could strengthen the recognition of local waveforms and interactions between variables.

The higher AUC but lower ACC on GunPoint and BasicMotions also motivates better probability calibration, threshold selection, and alignment of multiclass scores. Meanwhile, the contrast between EasyTPP and stock movement prediction highlights the need for training that explicitly targets future decisions. Future models should improve the reliability of these decisions while retaining direct, structured outputs.

\section{Conclusion}

This study evaluates Jev 1.13.0 as a Structured System 1 model for time-series decisions through a unified evaluation framework. The results reveal uneven performance: Jev recognizes selected explicit temporal patterns and achieves relatively consistent discrimination in event type prediction, while performance on many numerical sequence classification tasks and stock movement prediction remains close to chance-level discrimination. Several datasets also show useful ranking information that does not translate into accurate final decisions. These findings highlight temporal representation, domain adaptation, and decision-score alignment as priorities for future Structured System 1 models. The framework supports extending this evaluation to additional models under consistent task definitions.

\bibliographystyle{plainnat}
\bibliography{references}

\clearpage
\appendix
\section{Complete UCR and UEA Results}
Tables~\ref{tab:ucr_binary_full}--\ref{tab:uea_full} present the complete
UCR and UEA classification results.
\begin{table}[!htbp]
\centering
\caption{Complete UCR binary results recorded in the source file. Missing scores are shown as --.}
\label{tab:ucr_binary_full}
\begin{minipage}[t]{0.485\textwidth}
\vspace{0pt}
\centering
\scriptsize
\setlength{\tabcolsep}{2pt}
\renewcommand{\arraystretch}{1.12}
\begin{tabular}{@{}p{0.68\linewidth}rr@{}}
\toprule
Dataset & ACC & AUC \\
\midrule
Beetle\allowbreak{}Fly & 0.4500 & 0.4200 \\
Bird\allowbreak{}Chicken & 0.5500 & 0.5800 \\
Chinatown & 0.6793 & 0.6657 \\
Coffee & 0.5357 & 0.5128 \\
Computers & 0.5120 & 0.5022 \\
Distal\allowbreak{}Phalanx\allowbreak{}Outline\allowbreak{}Correct & 0.5688 & 0.5898 \\
Dodger\allowbreak{}Loop\allowbreak{}Game & 0.5942 & 0.6475 \\
Dodger\allowbreak{}Loop\allowbreak{}Weekend & 0.7464 & 0.6953 \\
Earthquakes & -- & -- \\
ECG200 & 0.5900 & 0.4696 \\
ECG\allowbreak{}Five\allowbreak{}Days & 0.4901 & 0.4434 \\
Ford\allowbreak{}A & 0.5114 & 0.5773 \\
Ford\allowbreak{}B & 0.5173 & 0.5324 \\
Freezer\allowbreak{}Regular\allowbreak{}Train & 0.5274 & 0.6013 \\
Freezer\allowbreak{}Small\allowbreak{}Train & 0.5246 & 0.5982 \\
Gun\allowbreak{}Point & 0.5200 & 0.7215 \\
Gun\allowbreak{}Point\allowbreak{}Age\allowbreak{}Span & 0.4747 & 0.4514 \\
Gun\allowbreak{}Point\allowbreak{}Male\allowbreak{}Versus\allowbreak{}Female & 0.4462 & 0.4655 \\
Gun\allowbreak{}Point\allowbreak{}Old\allowbreak{}Versus\allowbreak{}Young & 0.5460 & 0.5269 \\
Ham & 0.4952 & 0.4858 \\
Hand\allowbreak{}Outlines & 0.6405 & 0.5805 \\
\bottomrule
\end{tabular}
\end{minipage}\hfill%
\begin{minipage}[t]{0.485\textwidth}
\vspace{0pt}
\centering
\scriptsize
\setlength{\tabcolsep}{2pt}
\renewcommand{\arraystretch}{1.12}
\begin{tabular}{@{}p{0.68\linewidth}rr@{}}
\toprule
Dataset & ACC & AUC \\
\midrule
Herring & 0.5469 & 0.4165 \\
House\allowbreak{}Twenty & 0.6134 & 0.5048 \\
Italy\allowbreak{}Power\allowbreak{}Demand & 0.5598 & 0.6597 \\
Lightning2 & 0.5574 & 0.3956 \\
Middle\allowbreak{}Phalanx\allowbreak{}Outline\allowbreak{}Correct & 0.5533 & 0.4993 \\
Mote\allowbreak{}Strain & 0.5391 & 0.6209 \\
Phalanges\allowbreak{}Outlines\allowbreak{}Correct & 0.6107 & 0.5219 \\
Power\allowbreak{}Cons & 0.5167 & 0.6019 \\
Proximal\allowbreak{}Phalanx\allowbreak{}Outline\allowbreak{}Correct & 0.6460 & 0.5277 \\
Semg\allowbreak{}Hand\allowbreak{}Gender\allowbreak{}Ch2 & 0.4683 & 0.5101 \\
Shapelet\allowbreak{}Sim & 0.4778 & 0.3522 \\
Sony\allowbreak{}AIBO\allowbreak{}Robot\allowbreak{}Surface1 & 0.5408 & 0.2929 \\
Sony\allowbreak{}AIBO\allowbreak{}Robot\allowbreak{}Surface2 & 0.3442 & 0.2387 \\
Strawberry & 0.4378 & 0.5830 \\
Toe\allowbreak{}Segmentation1 & 0.6009 & 0.5998 \\
Toe\allowbreak{}Segmentation2 & 0.6308 & 0.6048 \\
Two\allowbreak{}Lead\allowbreak{}ECG & 0.5004 & 0.4629 \\
Wafer & 0.6525 & 0.8056 \\
Wine & 0.4815 & 0.5700 \\
Worms\allowbreak{}Two\allowbreak{}Class & 0.4935 & 0.5492 \\
Yoga & 0.4947 & 0.5128 \\
\bottomrule
\end{tabular}
\end{minipage}
\end{table}

\begin{table}[!htbp]
\centering
\caption{UCR results.}
\label{tab:ucr_multiclass_full}
\begin{minipage}[t]{0.485\textwidth}
\vspace{0pt}
\centering
\scriptsize
\setlength{\tabcolsep}{2pt}
\renewcommand{\arraystretch}{1.12}
\begin{tabular}{@{}p{0.68\linewidth}rr@{}}
\toprule
Dataset & ACC & AUC \\
\midrule
ACSF1 & 0.1000 & 0.4420 \\
Adiac & 0.0179 & 0.4969 \\
All\allowbreak{}Gesture\allowbreak{}Wiimote\allowbreak{}X & 0.1443 & 0.5621 \\
All\allowbreak{}Gesture\allowbreak{}Wiimote\allowbreak{}Y & 0.1057 & 0.4572 \\
All\allowbreak{}Gesture\allowbreak{}Wiimote\allowbreak{}Z & 0.1171 & 0.5530 \\
Arrow\allowbreak{}Head & 0.3429 & 0.4816 \\
Beef & 0.2000 & 0.5597 \\
BME & 0.3933 & 0.7434 \\
Car & 0.2833 & 0.5857 \\
CBF & 0.3889 & 0.7019 \\
Chlorine\allowbreak{}Concentration & 0.2698 & 0.4654 \\
Cin\allowbreak{}CECG\allowbreak{}Torso & 0.2536 & 0.4655 \\
Cricket\allowbreak{}X & 0.0795 & 0.4812 \\
Cricket\allowbreak{}Y & 0.0718 & 0.4916 \\
Cricket\allowbreak{}Z & 0.0821 & 0.4945 \\
Crop & 0.0475 & 0.5273 \\
Diatom\allowbreak{}Size\allowbreak{}Reduction & 0.2876 & 0.5240 \\
Distal\allowbreak{}Phalanx\allowbreak{}Outline\allowbreak{}Age\allowbreak{}Group & 0.3813 & 0.4547 \\
Distal\allowbreak{}Phalanx\allowbreak{}TW & 0.1007 & 0.4074 \\
Dodger\allowbreak{}Loop\allowbreak{}Day & 0.2125 & 0.5056 \\
ECG5000 & 0.3811 & 0.5590 \\
Electric\allowbreak{}Devices & 0.1323 & 0.5146 \\
EOG\allowbreak{}Horizontal\allowbreak{}Signal & 0.0718 & 0.5048 \\
EOG\allowbreak{}Vertical\allowbreak{}Signal & 0.0718 & 0.4936 \\
Ethanol\allowbreak{}Level & 0.2520 & 0.5199 \\
Face\allowbreak{}All & 0.0686 & 0.5063 \\
Face\allowbreak{}Four & 0.2500 & 0.5375 \\
Faces\allowbreak{}UCR & 0.0683 & 0.5190 \\
Fifty\allowbreak{}Words & 0.0835 & 0.5048 \\
Fish & 0.1543 & 0.5235 \\
Fungi & 0.0645 & 0.5175 \\
Gesture\allowbreak{}Mid\allowbreak{}Air\allowbreak{}D1 & 0.0692 & 0.5705 \\
Gesture\allowbreak{}Mid\allowbreak{}Air\allowbreak{}D2 & 0.0846 & 0.5714 \\
Gesture\allowbreak{}Mid\allowbreak{}Air\allowbreak{}D3 & 0.0385 & 0.5762 \\
Gesture\allowbreak{}Pebble\allowbreak{}Z1 & 0.1744 & 0.4911 \\
Gesture\allowbreak{}Pebble\allowbreak{}Z2 & 0.1329 & 0.4943 \\
Haptics & 0.2143 & 0.5249 \\
Inline\allowbreak{}Skate & 0.0964 & 0.4832 \\
Insect\allowbreak{}EPG\allowbreak{}Regular\allowbreak{}Train & 0.3373 & 0.5308 \\
Insect\allowbreak{}EPG\allowbreak{}Small\allowbreak{}Train & 0.3373 & 0.5304 \\
Insect\allowbreak{}Wingbeat\allowbreak{}Sound & 0.0975 & 0.5110 \\
Large\allowbreak{}Kitchen\allowbreak{}Appliances & 0.2907 & 0.3956 \\
Lightning7 & 0.1233 & 0.4915 \\
\bottomrule
\end{tabular}
\end{minipage}\hfill%
\begin{minipage}[t]{0.485\textwidth}
\vspace{0pt}
\centering
\scriptsize
\setlength{\tabcolsep}{2pt}
\renewcommand{\arraystretch}{1.12}
\begin{tabular}{@{}p{0.68\linewidth}rr@{}}
\toprule
Dataset & ACC & AUC \\
\midrule
Mallat & 0.1190 & 0.5306 \\
Meat & 0.3333 & 0.5294 \\
Medical\allowbreak{}Images & 0.1000 & 0.4920 \\
Melbourne\allowbreak{}Pedestrian & 0.1095 & 0.5190 \\
Middle\allowbreak{}Phalanx\allowbreak{}Outline\allowbreak{}Age\allowbreak{}Group & 0.4545 & 0.4114 \\
Middle\allowbreak{}Phalanx\allowbreak{}TW & 0.0974 & 0.4648 \\
Mixed\allowbreak{}Shapes\allowbreak{}Regular\allowbreak{}Train & 0.2256 & 0.5093 \\
Mixed\allowbreak{}Shapes\allowbreak{}Small\allowbreak{}Train & 0.2243 & 0.5029 \\
Non\allowbreak{}Invasive\allowbreak{}Fetal\allowbreak{}ECG\allowbreak{}Thorax1 & 0.0204 & 0.4941 \\
Non\allowbreak{}Invasive\allowbreak{}Fetal\allowbreak{}ECG\allowbreak{}Thorax2 & 0.0183 & 0.5045 \\
Olive\allowbreak{}Oil & 0.2667 & 0.5510 \\
OSU\allowbreak{}Leaf & 0.2149 & 0.5213 \\
Phoneme & 0.0211 & 0.5020 \\
Pickup\allowbreak{}Gesture\allowbreak{}Wiimote\allowbreak{}Z & 0.1200 & 0.5004 \\
Pig\allowbreak{}Airway\allowbreak{}Pressure & 0.0192 & 0.4898 \\
Pig\allowbreak{}Art\allowbreak{}Pressure & 0.0192 & 0.4996 \\
Pig\allowbreak{}CVP & 0.0192 & 0.4978 \\
PLAID & 0.0652 & 0.5016 \\
Plane & 0.1048 & 0.5274 \\
Proximal\allowbreak{}Phalanx\allowbreak{}Outline\allowbreak{}Age\allowbreak{}Group & 0.3902 & 0.4332 \\
Proximal\allowbreak{}Phalanx\allowbreak{}TW & 0.0195 & 0.3344 \\
Refrigeration\allowbreak{}Devices & 0.3547 & 0.5161 \\
Rock & 0.4200 & 0.5535 \\
Screen\allowbreak{}Type & 0.3333 & 0.4908 \\
Semg\allowbreak{}Hand\allowbreak{}Movement\allowbreak{}Ch2 & 0.1578 & 0.4778 \\
Semg\allowbreak{}Hand\allowbreak{}Subject\allowbreak{}Ch2 & 0.1822 & 0.5032 \\
Shake\allowbreak{}Gesture\allowbreak{}Wiimote\allowbreak{}Z & 0.1000 & 0.4560 \\
Shapes\allowbreak{}All & 0.0167 & 0.5100 \\
Small\allowbreak{}Kitchen\allowbreak{}Appliances & 0.3280 & 0.4814 \\
Smooth\allowbreak{}Subspace & 0.9000 & 0.9815 \\
Star\allowbreak{}Light\allowbreak{}Curves & 0.2084 & 0.5856 \\
Swedish\allowbreak{}Leaf & 0.0800 & 0.5429 \\
Symbols & 0.1678 & 0.5365 \\
Synthetic\allowbreak{}Control & 0.8100 & 0.9978 \\
Trace & 0.2100 & 0.5328 \\
Two\allowbreak{}Patterns & 0.1380 & 0.3224 \\
UMD & 0.4306 & 0.6121 \\
U\allowbreak{}Wave\allowbreak{}Gesture\allowbreak{}Library\allowbreak{}All & 0.1309 & 0.4791 \\
U\allowbreak{}Wave\allowbreak{}Gesture\allowbreak{}Library\allowbreak{}X & 0.1259 & 0.4971 \\
U\allowbreak{}Wave\allowbreak{}Gesture\allowbreak{}Library\allowbreak{}Y & 0.1262 & 0.4968 \\
U\allowbreak{}Wave\allowbreak{}Gesture\allowbreak{}Library\allowbreak{}Z & 0.1265 & 0.4859 \\
Word\allowbreak{}Synonyms & 0.0502 & 0.4875 \\
Worms & 0.2857 & 0.4766 \\
\bottomrule
\end{tabular}
\end{minipage}
\end{table}

\begin{table}[!htbp]
\centering
\caption{UEA results}
\label{tab:uea_full}
\begin{minipage}[t]{0.485\textwidth}
\vspace{0pt}
\centering
\scriptsize
\setlength{\tabcolsep}{2pt}
\renewcommand{\arraystretch}{1.12}
\begin{tabular}{@{}p{0.68\linewidth}rr@{}}
\toprule
Dataset & ACC & AUC \\
\midrule
Articulary\allowbreak{}Word\allowbreak{}Recognition & 0.0400 & 0.4802 \\
Atrial\allowbreak{}Fibrillation & 0.3333 & 0.3567 \\
Basic\allowbreak{}Motions & 0.3000 & 0.7671 \\
Character\allowbreak{}Trajectories & 0.0599 & 0.5156 \\
Epilepsy & 0.2971 & 0.6622 \\
E\allowbreak{}Ring & 0.1667 & 0.5620 \\
Finger\allowbreak{}Movements & 0.4400 & 0.4138 \\
Handwriting & 0.0412 & 0.4986 \\
Japanese\allowbreak{}Vowels & 0.1243 & 0.4935 \\
\bottomrule
\end{tabular}
\end{minipage}\hfill%
\begin{minipage}[t]{0.485\textwidth}
\vspace{0pt}
\centering
\scriptsize
\setlength{\tabcolsep}{2pt}
\renewcommand{\arraystretch}{1.12}
\begin{tabular}{@{}p{0.68\linewidth}rr@{}}
\toprule
Dataset & ACC & AUC \\
\midrule
Libras & 0.0667 & 0.4950 \\
LSST & 0.0243 & 0.4898 \\
NATOPS & 0.1222 & 0.4552 \\
Pen\allowbreak{}Digits & 0.1335 & 0.5998 \\
Phoneme\allowbreak{}Spectra & 0.0316 & 0.5160 \\
Racket\allowbreak{}Sports & 0.2566 & 0.4613 \\
Spoken\allowbreak{}Arabic\allowbreak{}Digits & 0.0991 & 0.5040 \\
U\allowbreak{}Wave\allowbreak{}Gesture\allowbreak{}Library & 0.1250 & 0.5217 \\
\bottomrule
\end{tabular}
\end{minipage}
\end{table}

\end{document}